%% file: main.tex
\documentclass[letterpaper]{article} 
\usepackage[preprint]{aaai2027}

\usepackage{amsthm}

\usepackage{amssymb}
\usepackage{amsfonts}
\usepackage{mathtools}
\usepackage[table]{xcolor}
\usepackage{tabularx}
\usepackage{array}

\theoremstyle{plain}

\theoremstyle{definition}

\theoremstyle{remark}

\usepackage{xcolor}
\usepackage{algorithm}
\usepackage{algorithmic}
\usepackage{booktabs} 

\usepackage{multirow}
\usepackage{enumitem}
\usepackage[hyphens]{url}  
\usepackage{graphicx} 
\usepackage{natbib}  
\usepackage{caption} 
\usepackage{makecell}
\usepackage{amsmath} 
\usepackage{algorithm}
\usepackage{algorithmic}

\newcommand{\ourmethod}{\textit{Lever-Edit}}

\usepackage{newfloat}
\usepackage{listings}
\DeclareCaptionStyle{ruled}{labelfont=normalfont,labelsep=colon,strut=off} 
\floatstyle{ruled}
\newfloat{listing}{tb}{lst}{}
\floatname{listing}{Listing}

\usepackage{booktabs}

\title{Can We Perform Online RL for Image Editing without Editing Rewards?}
\author{
    Qichao Ma\textsuperscript{\rm 1}\equalcontrib,
    Jikang Cheng\textsuperscript{\rm 1}\equalcontrib\thanks{Project Lead.},
    Ling Liang\textsuperscript{\rm 1},
    Zhaofei Yu\textsuperscript{\rm 1},
    Tiejun Huang\textsuperscript{\rm 1},
    Renye Yan\textsuperscript{\rm 1}\corresponding
}
\affiliations{
    \textsuperscript{\rm 1}Peking University
}

\begin{document}

\maketitle
\begin{abstract}
Reinforcement learning (RL) enables direct preference optimization for image editing through editing-specific rewards, which remain less developed due to costly triplet supervision and complex task-dependent calibration. In contrast, text-to-image (T2I) generation benefits from a mature and diverse reward ecosystem spanning semantic alignment, aesthetics, realism, glyph shape, and other visual preferences. Extending this ecosystem to image editing would substantially broaden the range of visual preferences accessible to RL-based optimization, prompting the central question: \emph{Can We Perform Image Editing RL without Editing Rewards?} 
In this paper, we argue that the standard image editing dimensions have potential to be mapped to the T2I reward space: image quality can transfer directly, prompt following can be aligned through a description of the desired visual state, and reference consistency admits a coarse semantic conversion by encoding the source content to preserve. 
However, editing instructions specify relative changes, whereas T2I rewards require self-contained target descriptions; moreover, semantically valid captions from generic vision-language models may be incompatible with the frozen reward. Hence, we further introduce \ourmethod{}, a two-stage framework that learns a reward-aligned captioner for counterfactual target descriptions, freezes it, and optimizes the editing policy solely with the transferred T2I reward. Experiments show competitive editing alignment and source preservation against editing-reward-based fine-tuning, while outperforming intuitive transfer baselines.
\end{abstract}

\input{chapter/Intro}
\input{chapter/Related_works}
\input{chapter/Problem_Formulation}
\input{chapter/Method}

\input{chapter/Experiments}

\input{chapter/Conclusion}

\bibliography{aaai2027}


\end{document}

%% file: chapter/Intro.tex
\section{Introduction}

\begin{figure}[t]
    \centering
    \includegraphics[width=\linewidth]{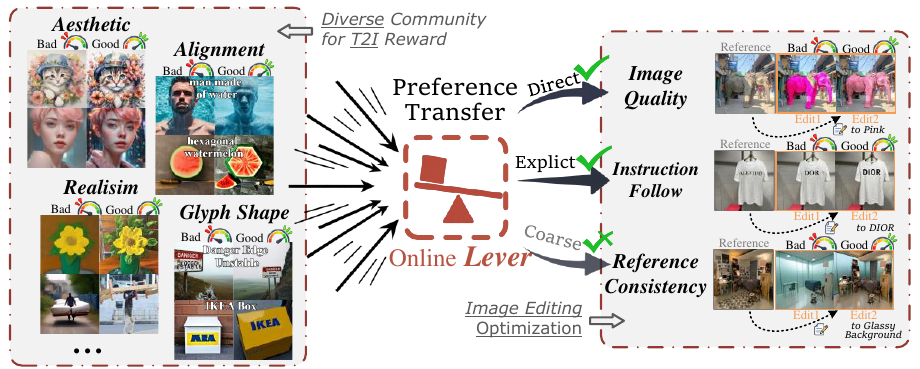}
    \caption{Motivation for T2I-to-editing reward transfer. \ourmethod{} acts as an online lever that harnesses the collective power of the broad T2I reward community, transferring its mature and diverse visual preferences into effective supervision for image-editing RL.}
    \label{fig:motivation}
\end{figure}

Instruction-guided image editing enables users to modify an existing image through natural-language instructions while preserving content unrelated to the requested change~\cite{huang2025diffusion}. Although recent diffusion and flow-based editors have achieved substantial progress~\cite{nichol2021glide,liu2025step1x,yan2026pixel}, supervised fine-tuning optimizes likelihood over curated editing pairs rather than task-level human preferences. Reinforcement learning (RL) instead directly optimizes scalar or rationale-based feedback on sampled results~\cite{black2023training,fan2023dpok,guo2026leveraging,liu2026unigrpo}. At the center of editing RL lies the reward. However, constructing a reliable editing reward requires joint reasoning over the reference image, editing instruction, and edited output, as well as costly triplet-level supervision. The desired balance between executing the instruction and preserving source content further varies across edit types, complicating reward calibration and generalization~\cite{wu2025editreward,luo2026editscore,zhao2026trust,long2026spatialreward}.
Editing quality is commonly characterized along three dimensions: \emph{pure image quality}, \emph{edit execution}, and \emph{reference consistency}~\cite{luo2026editscore,zhao2026trust}. Pure image quality concerns intrinsic properties of the edited output, edit execution measures whether the requested transformation is realized, and reference consistency measures the preservation of source content unrelated to the edit. Surprisingly, we find that all three dimensions exhibit potential mappings to the T2I reward space: image quality can be evaluated directly, edit execution can be reformulated as alignment with a description of the desired visual state, and reference consistency can be coarsely projected by incorporating preserved source semantics into that description, as illustrated in Fig.~\ref{fig:motivation}.

This raises a central question: \emph{Can We Perform Image Editing RL without Editing Rewards}, instead using t2i reward? Such a formulation offers two advantages. First, it gives editing RL direct access to the mature and diverse reward ecosystem of image generation, including rewards for semantic alignment, aesthetics, realism, OCR, safety, and other visual preferences~\cite{xu2023imagereward,wu2023human,kirstain2023pick,ye2025realgen,zhu2026textpecker,wang2025unifiedreward,huang2026readitback,lamba2025alignment}. Improvements in these rewards can consequently be transferred to editing without redesigning the alignment objective. Second, it removes the dependence on costly editing-evaluator development and deployment.

Directly applying a T2I reward to image editing is, however, rather challenging. First, the two tasks employ fundamentally different conditioning semantics. An editing instruction defines a relative transformation with respect to a reference image, whereas a T2I reward expects a self-contained description of the target image. Directly using the instruction preserves the requested change but omits the source semantics that should remain unchanged, necessitating a reference-aware conversion into a counterfactual target description. Second, a semantically valid target description is not necessarily compatible with the frozen T2I reward. An off-the-shelf Vision-Language Model (VLM) may produce a plausible caption whose representation is poorly calibrated to the reward model's evaluation space, thereby weakening the resulting optimization signal. Therefore, effective reward transfer requires both edit-target conversion and explicit reward alignment, which remains a challenging issue.

In this paper, we introduce \ourmethod{}, a reward-transfer framework in which a lightweight reward-aligned captioner acts as a \emph{lever}: it enables a mature T2I reward to provide effective supervision for image-editing RL without constructing or invoking an editing reward. Specifically, \ourmethod{} follows a two-stage design. In the first stage, we adapt a VLM through prefix tuning to map a reference image and an editing instruction to a counterfactual target caption. The caption jointly describes the intended modification and the source semantics that should be retained. We further align the captioner's representation with the frozen T2I reward, producing reward-compatible conditions rather than generic image descriptions. In the second stage, we freeze the captioner and optimize the diffusion editing policy using only the transferred T2I reward. No editing reward is used during policy fine-tuning. This design separates reward transfer from policy optimization and can be applied as a plug-in to different VLM and image-editing backbones.

Our contributions are summarized as follows:
\begin{itemize}[leftmargin=*]
    \item We revisit the standard dimensions of editing quality from the perspective of reward transfer and identify their potential correspondence to general image evaluation, motivating image-editing RL without an editing-specific reward.
    \item We propose \ourmethod{}, a two-stage framework that converts reference-conditioned editing objectives into reward-aligned target captions and fine-tunes an editing policy using only a frozen T2I reward.
    \item We empirically demonstrate that transferred T2I rewards can provide competitive editing alignment and source preservation without an image-to-image reward model during policy fine-tuning.
\end{itemize}

%% file: chapter/Related_works.tex
\section{Related Works}

\subsection{Reinforcement Learning for Image Editing}

Reinforcement learning enables direct preference optimization for visual generation without dense task-specific annotations. DDPO~\cite{black2023training} and DPOK~\cite{fan2023dpok} pioneered RL-based fine-tuning for text-to-image diffusion models. DanceGRPO~\cite{xue2025dancegrpo} and Flow-GRPO~\cite{liu2025flow} subsequently broadened this paradigm to contemporary diffusion and flow-matching backbones, establishing online RL as a general post-training framework for visual generation. Further studies improve optimization efficiency and stability through forward-process policy updates, pretraining-aligned objectives, entropy-adaptive exploration, and selective intervention along the denoising trajectory~\cite{zheng2026diffusionnft,xue2026awm,yan2025entropy,yan2026less}. RL paradigms developed for T2I generation have subsequently been successfully adapted to preference optimization for instruction-guided image editing. In this setting, RL aims to further improve prompt adherence, reference consistency, and overall visual quality, with recent methods demonstrating encouraging gains along these dimensions~\cite{luo2026editscore,zhao2026trust,guo2026leveraging}.

\subsection{Reward Modeling in Image RL}

\textbf{Image Generation.} Reward modeling for text-to-image generation has evolved into a diverse ecosystem spanning general human preferences and task-specific criteria. ImageReward~\cite{xu2023imagereward}, HPS~\cite{wu2023human,hpsv2}, and PickScore~\cite{kirstain2023pick} capture broad judgments of semantic alignment and visual appeal. Specialized rewards target photorealism in RealGen~\cite{ye2025realgen} and visual-text fidelity in TextPecker~\cite{zhu2026textpecker}, while UnifiedReward~\cite{wang2025unifiedreward} supports pointwise and pairwise multimodal evaluation and SigLIP~\cite{zhai2023siglip} provides general text--image correspondence scores. Complementing explicitly trained evaluators, SpectraReward~\cite{huang2026readitback} repurposes pretrained MLLMs as training-free, zero-shot rewards through image-conditioned prompt likelihood. This breadth makes the T2I reward ecosystem a compelling source of supervision for broader visual alignment tasks.\\
\textbf{Image Editing.} Editing rewards evaluate the triplet $(I_{\mathrm{ref}}, q_{\mathrm{edit}}, I_{\mathrm{edit}})$ across output quality, edit execution, and reference consistency. EditReward~\cite{wu2025editreward} constructs over 200K expert-annotated preference pairs and trains a multidimensional uncertainty-aware ranker. EditScore~\cite{luo2026editscore} follows a pipeline from benchmark construction and reward-data curation to reward-model training, self-ensemble, and online RL. FIRM Reward~\cite{zhao2026trust} employs a difference-first MLLM pipeline, specialized reward-model training, and consistency-modulated reward fusion. SpatialReward~\cite{long2026spatialreward} further introduces a multi-stage spatial-grounding and verification pipeline, followed by progressive SFT and GRPO training. Such staged construction is a common consequence of the intrinsic difficulty of evaluating relative edits and entails substantial design and implementation costs. Moreover, these rewards remain largely centered on edit execution and reference consistency, with limited coverage of the broader alignment preferences supported by T2I rewards. This restricted reward space constrains the optimization potential of image-editing RL. 

%% file: chapter/Problem_Formulation.tex
\section{Problem Formulation}

\subsection{The Reward-Transfer Gap}

Let $I_{\mathrm{ref}}$ denote a reference image, $q_{\mathrm{edit}}$ an editing instruction, and $I_{\mathrm{edit}}$ the output sampled from an editing policy:
\begin{equation}
    I_{\mathrm{edit}} \sim \pi_\theta(\cdot \mid I_{\mathrm{ref}}, q_{\mathrm{edit}}).
\end{equation}
Conventional image-editing RL optimizes a task-specific reward defined on the complete editing triplet,
\begin{equation}
    \max_\theta\; \mathbb{E}\!\left[
    R_{\mathrm{edit}}(I_{\mathrm{ref}},q_{\mathrm{edit}},I_{\mathrm{edit}})
    \right].
    \label{eq:editing_reward_objective}
\end{equation}
The reference image allows $R_{\mathrm{edit}}$ to determine both what should change and what should remain unchanged. By contrast, a T2I reward follows an image--text interface,
\begin{equation}
    R_{\mathrm{T2I}}(I,c),
\end{equation}
where $c$ is expected to be a self-contained description of the visual content in $I$. Directly substituting $q_{\mathrm{edit}}$ for $c$ therefore creates a fundamental conditioning mismatch. An editing instruction specifies a relative transformation, such as ``make the car red,'' whose target state depends on the particular reference image. It neither describes the complete target image nor exposes the source semantics that should be preserved. Consequently, $R_{\mathrm{T2I}}(I_{\mathrm{edit}},q_{\mathrm{edit}})$ may reward the requested attribute while remaining insensitive to the loss of reference-specific content.

Reward transfer thus requires a condition converter
$g:\mathcal I\times\mathcal Q\rightarrow\mathcal C$ that maps a reference image and a relative editing instruction to a self-contained target description:
\begin{equation}
    c_{\mathrm{tgt}} = g(I_{\mathrm{ref}},q_{\mathrm{edit}}).
    \label{eq:reward_transfer}
\end{equation}
The edited output is then evaluated by $R_{\mathrm{T2I}}(I_{\mathrm{edit}},c_{\mathrm{tgt}})$. The central problem is whether $g$ can retain the information required for editing evaluation while producing a condition that remains effective for a frozen T2I reward.

\subsection{Mapping Editing Quality to the T2I Reward Space}

Editing quality is commonly assessed along three dimensions: pure image quality, edit execution, and reference consistency~\cite{luo2026editscore,zhao2026trust}. We represent these criteria conceptually as:
\begin{equation}
    \mathbf r_{\mathrm{edit}}=
    \begin{bmatrix}
        r_{\mathrm{qual}}(I_{\mathrm{edit}}) \\
        r_{\mathrm{exec}}(I_{\mathrm{ref}},q_{\mathrm{edit}},I_{\mathrm{edit}}) \\
        r_{\mathrm{cons}}(I_{\mathrm{ref}},q_{\mathrm{edit}},I_{\mathrm{edit}})
    \end{bmatrix},
    \label{eq:editing_dimensions}
\end{equation}
without assuming a particular scalarization of the three criteria. Although their native formulation is reference-conditioned, each criterion admits a correspondence to the T2I reward space.

Among the three dimensions, pure image quality admits an immediate transfer: it is intrinsic to $I_{\mathrm{edit}}$ and can be assessed without reference to either the source image or the editing instruction. The apparent difficulty instead lies in edit execution and reference consistency, both of which are conventionally defined relative to $I_{\mathrm{ref}}$. For edit execution, however, this dependence arises primarily from the form of the condition. The instruction $q_{\mathrm{edit}}$ describes a relative change, whereas the evaluation objective concerns whether the corresponding visual state appears in the result. Once this intended outcome is expressed as an absolute post-edit description, edit execution becomes an image--text alignment problem of the form already addressed by T2I rewards.
Reference consistency appears to present the greatest obstacle, since it explicitly concerns the relation between $I_{\mathrm{ref}}$ and $I_{\mathrm{edit}}$. Nevertheless, the source semantics that should remain unchanged can be incorporated into the target description together with the requested modification. The target condition can therefore be viewed as the composition
\begin{equation}
    c_{\mathrm{tgt}} = c_{\mathrm{change}} \oplus c_{\mathrm{preserve}},
    \label{eq:target_composition}
\end{equation}
where $c_{\mathrm{change}}$ specifies the intended changes and $c_{\mathrm{preserve}}$ encodes the source semantics to preserve. This formulation recasts edit execution and reference consistency as alignment between $I_{\mathrm{edit}}$ and a unified target description, enabling their joint evaluation by a T2I reward.

This mapping should be understood as a semantic projection rather than an equivalence between T2I and image-pair evaluation. A textual condition cannot fully encode low-level identity or pixel-accurate preservation. It can, however, represent the principal semantics underlying image quality, edit execution, and reference consistency. Thus, although direct reward transfer is ill-posed at the interface level, the core editing objectives remain transferable in principle. The practical challenge is to construct a target condition that faithfully realizes this correspondence.

\subsection{Intuitive Reward-Transfer Protocols}

Equation~\ref{eq:reward_transfer} suggests several straightforward transfer protocols. We introduce three representative variants that differ in how they instantiate $g$.

\paragraph{Direct Edit Prompt.}
The simplest protocol directly sets $c_{\mathrm{tgt}}=q_{\mathrm{edit}}$. It incurs no conversion cost and retains the requested transformation, but treats a relative instruction as an absolute image description. The resulting reward condition omits reference-specific context and provides no explicit account of the content to be preserved.

\paragraph{Online VLM Rewriting.}
A general-purpose VLM can instead generate the target condition on demand,
\begin{equation}
    c_{\mathrm{tgt}}=g_{\mathrm{VLM}}(I_{\mathrm{ref}},q_{\mathrm{edit}}).
\end{equation}
This protocol introduces reference awareness and can express the intended post-edit state more completely. However, repeatedly invoking a VLM during reward computation adds inference overhead, while its generic captions are produced for general semantic description rather than calibrated to the evaluation behavior of the selected T2I reward.

\paragraph{Offline Proprietary-MLLM Rewriting.}
A strong proprietary MLLM may be used to precompute target descriptions offline. Owing to its instruction-following and visual-reasoning capabilities, the resulting descriptions may accurately express the intended edit and preserved content from a human-semantic perspective. However, offline rewriting fixes these conditions independently of the downstream T2I reward, making them inflexible to the choice of reward model or subsequent optimization requirements. It also introduces dependence on a closed model. More fundamentally, semantic quality for human readers does not guarantee reward compatibility: an accurate description may emphasize attributes represented weakly by the reward or fall outside its preferred conditioning distribution.

Collectively, these protocols identify two requirements for effective reward transfer: the target description must faithfully encode both the requested modification and the source semantics to be preserved, and it must remain compatible with the frozen T2I reward. Direct use of the editing instruction fails to specify a complete target state, whereas generic VLM- or MLLM-based rewriting provides no guarantee of reward-specific alignment. These observations motivate the reward-aligned conversion mechanism presented next.

%% file: chapter/Method.tex
\section{Proposed Method}

In this section, we introduce the overall pipeline and method design of \ourmethod{}. \ourmethod{} consists of two stages. \textbf{Stage 1} addresses the mismatch in textual representations when incorporating T2I rewards. Specifically, we jointly optimize a learnable prefix together with a fixed system prompt to align with multiple T2I reward signals at both the embedding and textual levels, therefore ensuring execution and consistency signal with explicit text, and quality signal with hidden embedding. \textbf{Stage 2} then integrates the learned alignment into image-editing optimization. On the one hand, we freeze the parameters learned in Stage 1, allowing the resulting module to serve as a plug-in that can be seamlessly incorporated into existing image-editing fine-tuning frameworks without modifying their original training objectives. On the other hand, we inject the embedding-level alignment signal into the image-editing trajectory as auxiliary supervision, providing complementary guidance to balance prompt following and edit consistency.

\begin{figure*}[t]
    \centering
    \includegraphics[width=\linewidth]{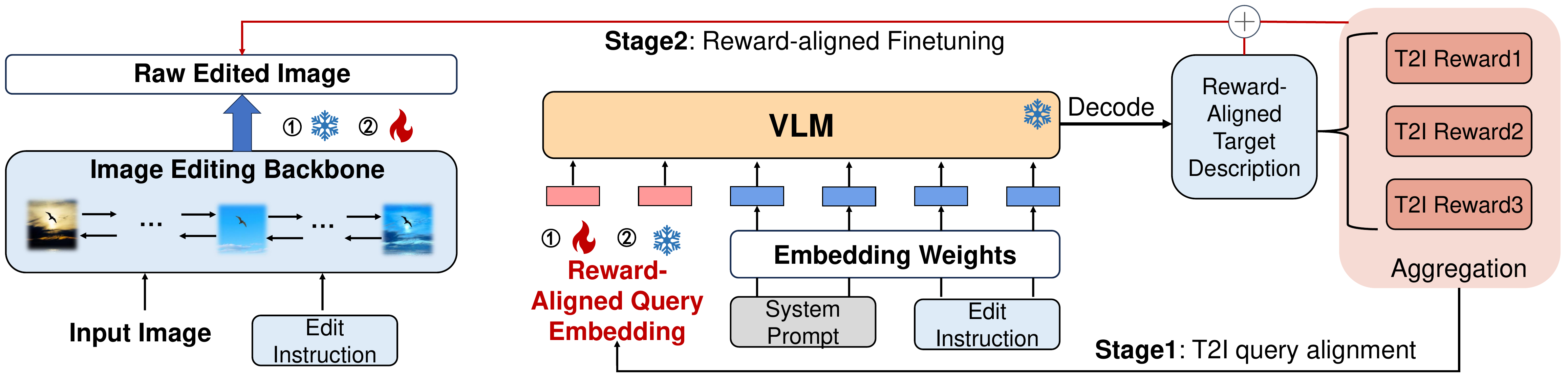}
    \caption{Overall two-stage T2I reward transfer pipeline in \ourmethod{}.}
    \label{fig:pipeline}
\end{figure*}
\subsection{T2I Query Alignment}

As illustrated on the right side of Fig.~\ref{fig:pipeline}, Stage 1 aims to align the VLM-generated target descriptions with the preferences of the downstream T2I reward models without updating either the image editing backbone or the VLM. To this end, we freeze both models and optimize only a set of learnable continuous query embeddings, denoted by $\mathcal{E}_{\mathrm{query}}$, through prefix tuning. Meanwhile, a fixed system prompt is explicitly concatenated with the original edit instruction to constrain the generation task at the semantic level. Conditioned on the reference image $I_{\mathrm{ref}}$, the explicit textual instruction $p_{\mathrm{sys}} \oplus q_{\mathrm{edit}}$, and the continuous edit query $\mathcal{E}_{\mathrm{query}}$, the VLM autoregressively generates a target description $c_{\mathrm{tgt}}$. Multiple complementary T2I reward models then evaluate the generated description, and their aggregated reward serves as the policy-gradient signal to optimize $\mathcal{E}_{\mathrm{query}}$. In this way, Stage~1 adapts the text-generation policy toward downstream T2I reward preferences while leaving the original model parameters unchanged.\\
\textbf{Semantic-Preserving Embedding-level Prefix Tuning.}
Reward-driven adaptation should improve downstream T2I compatibility without
altering the underlying generation task. Given a reference image
$I_{\mathrm{ref}}$ and its edit instruction $C_{\mathrm{edit}}$, we explicitly
concatenate a fixed system prompt $P_{\mathrm{sys}}$ with the edit instruction,
\begin{equation}
c_{\mathrm{in}}
=
[p_{\mathrm{sys}}  \oplus q_{\mathrm{edit}}],
\label{eq:stage1_text_input}
\end{equation}
where $ \oplus$ denotes concatenation along the token dimension. The system prompt explicitly instructs the VLM to describe the hypothetical edited image according to $I_{\mathrm{ref}}$ and $q_{\mathrm{edit}}$. It therefore provides a fixed, discrete specification of the target-description task throughout reward optimization.

In addition to the explicit textual instruction, we introduce a learnable continuous prefix
\begin{equation}
\begin{aligned}
\mathcal{E}_{\mathrm{query}}
=
[\mathcal{E}_1,\ldots,\mathcal{E}_{N_q}]
\in
\mathbb{R}^{N_q \times d},
\end{aligned}
\label{eq:edit_query}
\end{equation}
where $N_q$ is the number of learnable query tokens and $d$ denotes the hidden dimension of the VLM. Let $\mathcal{E}_{\mathrm{VLM}}^{text}(\cdot)$ denote the final text-domain input embedding of the VLM. The input is therefore constructed as
\begin{equation}
\begin{aligned}
\mathcal{E}_{VLM}^{text}
=
\Big[
\mathcal{E}_{\mathrm{query}}\,;\,
\mathcal{E}_{\mathrm{text}}\big(
[p_{\mathrm{sys}}
 \oplus q_{\mathrm{edit}}]
\big)
\Big],
\end{aligned}
\label{eq:prefix_input}
\end{equation}

Thus, $p_{\mathrm{sys}}  \oplus q_{\mathrm{edit}}$ explicitly specifies the
generation task in the discrete text space, whereas $q_{\mathrm{edit}}$
provides trainable degrees of freedom in the continuous embedding space.
Throughout Stage~1, all original VLM parameters remain frozen and only
$q_{\mathrm{edit}}$ is optimized. Compared with directly fine-tuning the VLM,
this design confines reward-driven adaptation to a lightweight prefix space,
thereby limiting interference with the pretrained language and semantic
capabilities while still providing sufficient flexibility to capture
T2I-reward-preferred generation patterns beyond a fixed handcrafted prompt.

For each sampled target description, we employ $K$ complementary T2I reward models to evaluate its compatibility with downstream image-generation objectives. Their outputs are aggregated as
\begin{equation}
R_{\mathrm{T2I}}
=
\sum_{k=1}^{K}
\omega_k
R_k
\left(
c_{\mathrm{tgt}};I_{\mathrm{ref}}
\right),
\label{eq:aggregated_reward}
\end{equation}
where $\omega_k$ denotes the weight of the $k$-th reward model.

We optimize $\mathcal{E}_{\mathrm{query}}$ to maximize the expected downstream T2I reward:
\begin{equation}
\begin{aligned}
J_{\mathrm{T2I}}
\left(
\mathcal{E}_{\mathrm{query}}
\right)
=
\mathbb{E}_{\substack{
(I_{\mathrm{ref}},q_{\mathrm{edit}})\sim\mathcal{D},\\
c_{\mathrm{tgt}}\sim\pi_{\mathcal{E}_{\mathrm{query}}}
}}
R_{\mathrm{T2I}},
\end{aligned}
\label{eq:t2i_reward_objective}
\end{equation}

Because $c_{\mathrm{tgt}}$ is obtained through discrete token sampling, the downstream reward cannot be directly differentiated through the generated sequence. We therefore optimize $q_{\mathrm{edit}}$ using policy gradient.\\
\textbf{Offline Token-level Distribution Alignment.}
Although the fixed system prompt explicitly specifies the target-description task reward-only optimization can still induce semantic drift with token-level mismatch. To provide an additional semantic anchor, we employ a strong offline MLLM (e.g., GPT or Gemini) to generate a high-quality reference description $c_{\mathrm{tgt}}^{*}$ for a small set of training samples.

Specifically, we construct an offline supervision set
\begin{equation}
\mathcal{D}_{\mathrm{gold}}
=
\left\{
\left(
I_{\mathrm{ref}}^{(i)},
q_{\mathrm{edit}}^{(i)},
c_{\mathrm{tgt}}^{*(i)}
\right)
\right\}_{i=1}^{N},
\label{eq:gold_dataset}
\end{equation}
where
$C_{\mathrm{tgt}}^{*}=(c_1^{*},\ldots,c_{L^{*}}^{*})$
denotes the reference description generated by the offline MLLM. At decoding step $t$, we represent the reference token $c_t^{*}$ using the empirical one-hot target distribution \begin{equation}
\mathcal{E}_{\mathrm{gold},t}(v)
=
\mathbb{I}
\left[
v=c_t^{*}
\right],
\qquad
v\in\mathcal{V},
\label{eq:gold_distribution}
\end{equation}
where $\mathcal{V}$ denotes the VLM vocabulary. For compactness, we use $\mathbb{E}_{\mathcal{D}_{\mathrm{gold}}}[\cdot]$ to denote expectation over samples from $\mathcal{D}_{\mathrm{gold}}$.

We regularize the query-conditioned policy toward this empirical semantic target through
\begin{equation}
\begin{aligned}
\mathcal{L}_{\mathrm{SFT}}
=
\mathbb{E}_{\mathcal{D}_{\mathrm{gold}}}
\Bigg[
\frac{1}{L^{*}}
\sum_{t=1}^{L^{*}}
D_{\mathrm{KL}}
\Big(
\mathcal{E}_{\mathrm{gold},t}
\,\Vert\,
\pi_{\mathcal{E}_{\mathrm{query}}}
\big(
\cdot
\mid I_{\mathrm{ref}}, \\[-0.2em]
p_{\mathrm{sys}} \oplus q_{\mathrm{edit}},
c_{\mathrm{tgt},<t}^{*}
\big)
\Big)
\Bigg].
\end{aligned}
\label{eq:sft_kl}
\end{equation}
Because $\mathcal{E}_{\mathrm{gold},t}$ is a one-hot distribution concentrated on $c_t^{*}$, the KL objective reduces exactly to the token-level negative log-likelihood used in standard SFT in eq. \ref{eq:sft_objective}.

\begin{equation}
\begin{aligned}
\mathcal{L}_{\mathrm{SFT}}
=
\mathbb{E}_{\mathcal{D}_{\mathrm{gold}}}
\Bigg[
\frac{1}{L^{*}}
\sum_{t=1}^{L^{*}}
\log
\pi_{\mathcal{E}_{\mathrm{query}}}
\Big(
c_t^{*}
\mid I_{\mathrm{ref}}, \\[-0.2em]
p_{\mathrm{sys}} \oplus q_{\mathrm{edit}},
c_{\mathrm{tgt},<t}^{*}
\Big)
\Bigg].
\end{aligned}
\label{eq:sft_objective}
\end{equation}

In all, the two objectives provide complementary supervision. Combining downstream reward alignment with offline semantic supervision, the overall objective of Stage 1 is
\begin{equation}
\boxed{
\mathcal{L}_{\mathrm{Stage1}}
=
\mathcal{L}_{\mathrm{PG}}
+
\lambda_{\mathrm{SFT}}
\mathcal{L}_{\mathrm{SFT}}
},
\label{eq:stage1_objective}
\end{equation} Policy gradient $\mathcal{L}_{\mathrm{PG}}$ increases the likelihood of description alignment that receives high downstream T2I rewards in embedding level, while $\mathcal{L}_{\mathrm{SFT}}$ regularizes the policy toward semantically reliable descriptions generated by the offline MLLM. The latter, therefore, mitigates semantic drift and potential hacking introduced by reward-driven optimization.

\subsection{Reward-aligned Finetuning}

Stage~1 freezes the image editing backbone and performs RL rollouts over the VLM-generated target descriptions $c_{\mathrm{tgt}}$, yielding a set of learnable query embeddings $\mathcal{E}_{\mathrm{query}}$ directly aligned with multiple T2I rewards. In Stage~2, the resulting query embeddings serve as a plug-and-play reward-alignment module that provides effective reward signals across different image-editing backbones and fine-tuning paradigms. Additionally, the corresponding VLM prefill embedding of $\mathcal{E}_{\mathrm{query}}$ has been contextualized with both visual and textual inputs. Therefore, additional reward-aligned multi-modal condition can be introduced into the image edit finetuning process.

%% file: chapter/Experiments.tex
\section{Experiments}

\begin{table*}[t]
\centering
\caption{Comparison of different methods across image editing evaluation metrics on Lever-Bench and GEdit \cite{liu2025step1x}.}
\label{tab:editing_metrics}

\footnotesize
\setlength{\tabcolsep}{3.0pt}
\renewcommand{\arraystretch}{1.08}

\begin{tabular}{@{}lccccccccc@{}}
\toprule

& \multicolumn{1}{c}{Consistency}
& \multicolumn{1}{c}{Aesthetics}
& \multicolumn{3}{c}{EditScore}
& \multicolumn{1}{c}{T2I Naturalness}
& \multicolumn{3}{c}{GEdit}
\\

\cmidrule(lr){2-2}
\cmidrule(lr){3-3}
\cmidrule(lr){4-6}
\cmidrule(lr){7-7}
\cmidrule(lr){8-10}

Method
& SigLIP
& HPSv3
& \makecell{Consistency}
& \makecell{Naturalness}
& \makecell{Execution}
& \makecell{RealGen v1}
& S
& Q
& O
\\

\midrule

w/o Alignment
& 0.0769
& 5.2742
& 0.8726
& 0.8811
& 0.6311
& 1.1958
& 6.181
& 6.330
& 5.763
\\

\textbf{Lever} Pure Edit
& 0.0772
& 5.9823
& 0.9044
& \textbf{\underline{0.8974}}
& \textbf{\underline{0.7307}}
& 1.2119
& 6.345
& 6.330
& 5.755
\\

\textbf{Lever} Pure VLM
& 0.0771
& 7.3152
& 0.9244
& 0.8837
& 0.6249
& 1.1946
& 6.058
& \textbf{\underline{6.661}}
& 5.706
\\

\textbf{Lever} Offline SOTA
& 0.0763
& 8.2645
& 0.9252
& 0.8896
& 0.7191
& 1.1995
& 6.042
& 6.389
& 5.517
\\

\textbf{\ourmethod{}}
& \textbf{\underline{0.0788}}
& \textbf{\underline{14.9762}}
& \textbf{\underline{0.9263}}
& 0.8822
& 0.7040
& \textbf{\underline{1.2532}}
& \textbf{\underline{6.364}}
& 6.362
& \textbf{\underline{5.788}}
\\

\bottomrule
\end{tabular}

\end{table*}

\subsection{Experiment Setup}
\textbf{Datasets.} We evaluate our method on both a failure-focused benchmark constructed for T2I-reward transfer and existing public image editing benchmarks. Specifically, we construct \textbf{Lever-Bench} to evaluate the transfer of T2I rewards for image editing. Rather than uniformly sampling common editing cases, we curate 1,880 image-instruction pairs $(I_{\mathrm{in}}, q_{\mathrm{edit}})$ from multiple sources, including Arena \cite{jiang2024genai}, Counting Edit \cite{geneval}, and EditScore-RL-Data \cite{luo2026editscore}, focusing on failure cases of existing image editing models. Lever-Bench spans four broad categories of editing scenarios: \emph{local edits}, including object removal, replacement, and material or attribute modification; \emph{human-centric edits}, including portrait, pose, action, and facial-expression changes; \emph{global edits}, such as background and style modification; and \emph{compositional edits} involving object counting and relational reasoning. To complement these targeted failure cases and evaluate generalization beyond our curated benchmark, we additionally report results on the open-source benchmarks: GEdit \cite{liu2025step1x} and EditReward-Bench \cite{luo2026editscore}.

\textbf{Rewards and Metrics.} Both training rewards and evaluation metrics are selected to cover three complementary dimensions of image editing quality: perceptual quality, edit execution, and reference consistency. For both Stage~1 and Stage~2, we employ HPSv3 \cite{ma2025hpsv3} and RealGen \cite{ye2025realgen} as T2I rewards. HPSv3 provides reward signals for edit execution, consistency, and aesthetics, while RealGen emphasizes realism and consistency. 
For evaluation, we separately assess the major quality dimensions to obtain a more comprehensive view of model performance. We measure perceptual quality from both aesthetic and realism perspectives. To evaluate semantic consistency, we use SigLIP \cite{zhai2023siglip} to measure the correspondence between the edited image and its textual condition. We further use EditScore \cite{luo2026editscore} to assess instruction execution and consistency between the input and edited images. These metrics jointly characterize perceptual quality, instruction following, and source consistency.

\textbf{Models and Baselines.} We instantiate \ourmethod{} on FLUX.1-Kontext-dev~\cite{labs2025flux} and apply
DiffusionNFT \cite{zheng2026diffusionnft}. 
Therefore, our baselines are designed to distinguish the effect of reward-aligned target description generation from the benefit of simply providing an alternative textual condition. In addition to the original image editing model without alignment, we compare four prompt constructions: (i) directly using the original edit instruction $q_{\mathrm{edit}}$ (\emph{\textbf{Lever} Pure Edit}); (ii) descriptions generated by the online VLM conditioned on the fixed $p_{\mathrm{sys}}$, without reward-driven query optimization (\emph{\textbf{Lever} Pure VLM}); (iii) high-quality descriptions generated offline by a strong closed-source MLLM (\emph{\textbf{Lever} Offline SOTA}); and (iv) reward-aligned descriptions generated by \emph{\textbf{\ourmethod{}}}. Unless otherwise stated, variants in comparison use the same image-editing backbone, edit instruction set, and training budget.

\begin{figure}[t]
    \centering
    \includegraphics[width=\linewidth]{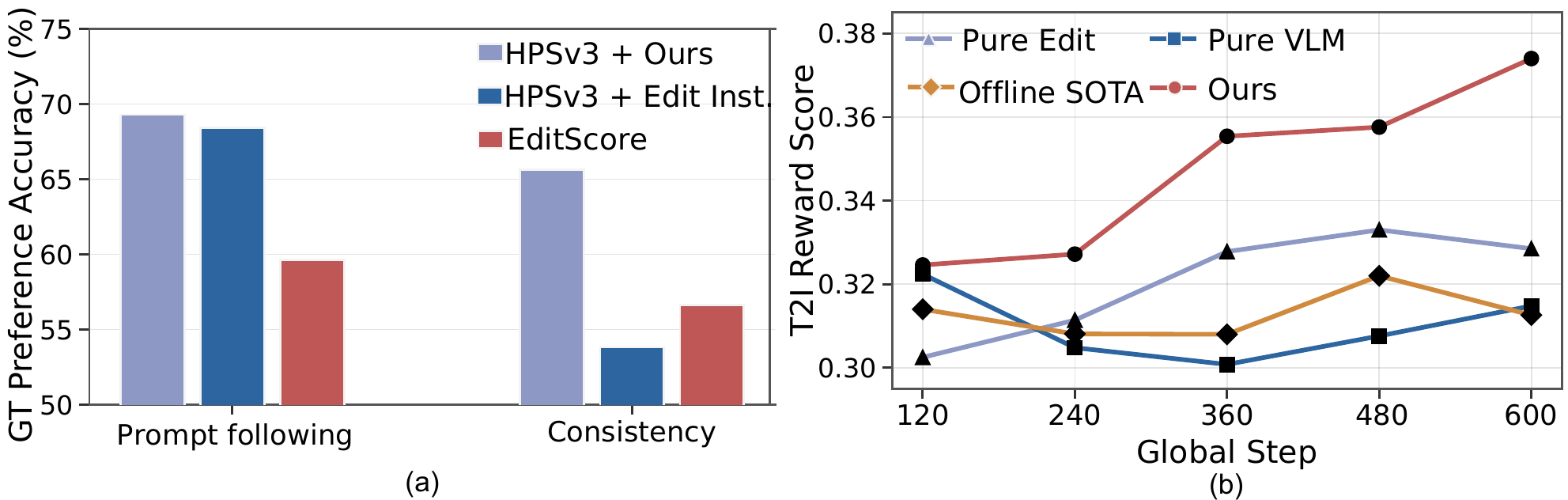}
    \caption{Effectiveness of T2I rewards transfer. (a) Evaluation of T2I reward and I2I reward (Editscore) on EditReward-Bench. (b) Evaluated T2I reward comparison of different methods. \ourmethod{} shows stable and best optimization.}
    \label{fig:T2I_effect}
\end{figure}

\begin{figure*}[htbp]
    \centering
    \includegraphics[width=\linewidth]{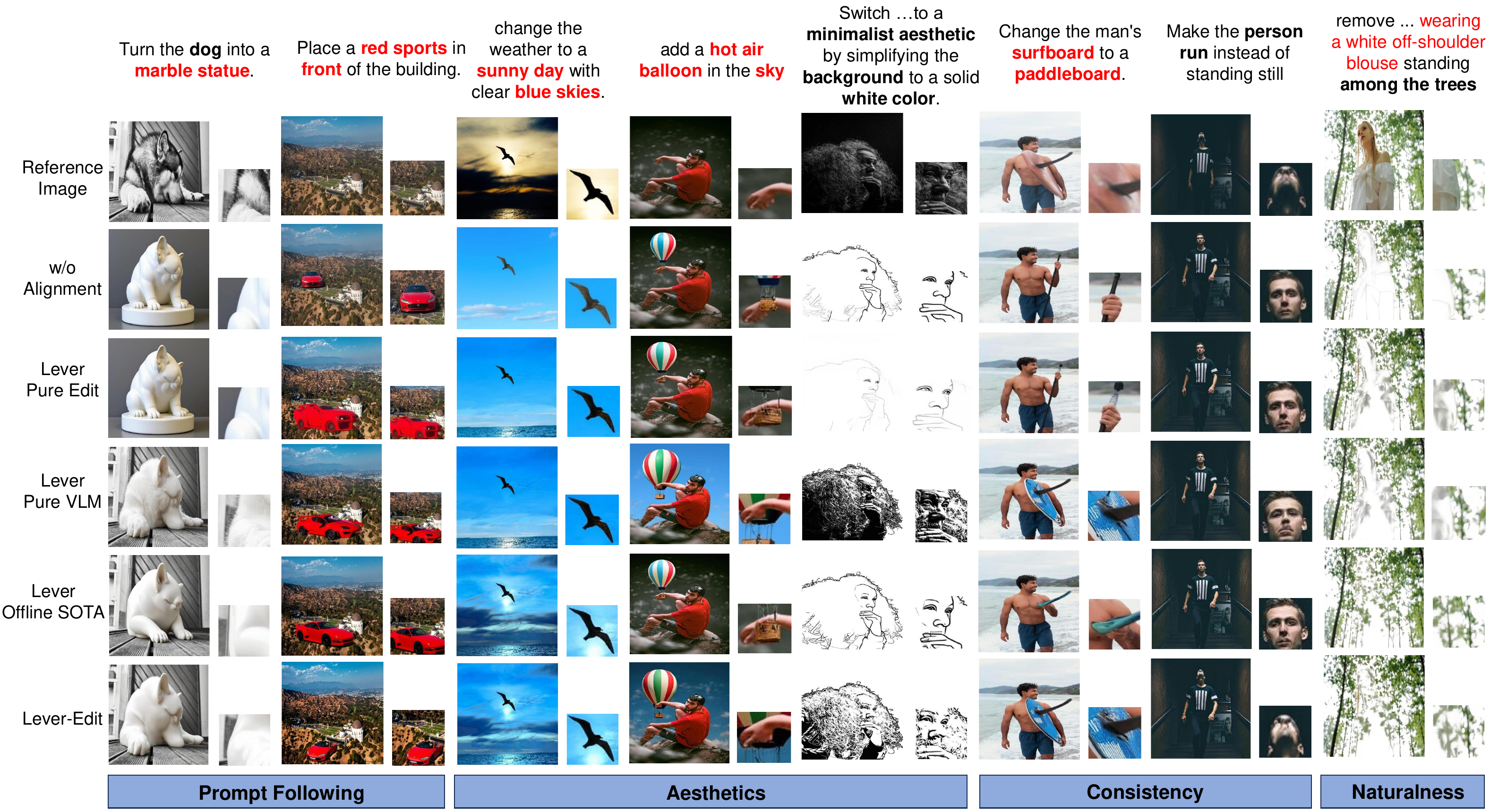}
    \caption{Qualitative analysis of different methods. \ourmethod{} successfully edits the images with strong prompt following and high aesthetics, while maintaining both semantic and source consistency with naturalness. Zoom in for better visualizations.}
    \label{fig:exp3}
\end{figure*}

\begin{figure*}[t]
    \centering
    \includegraphics[width=\linewidth]{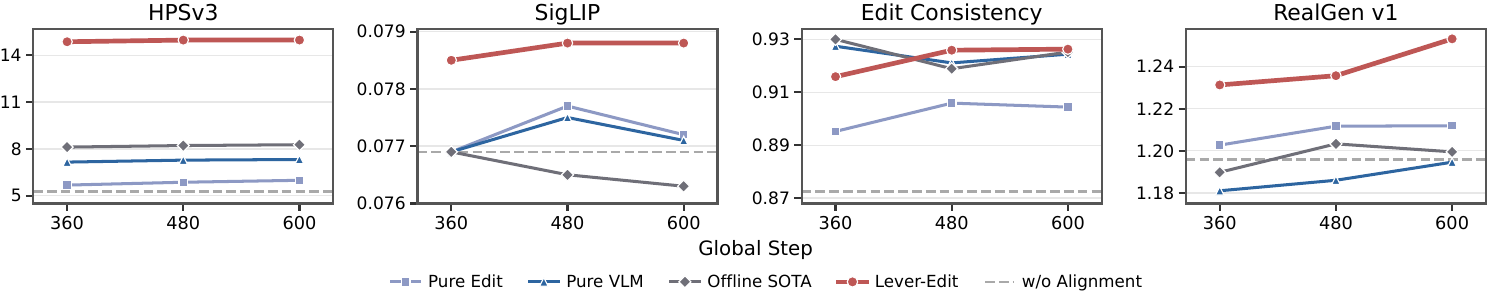}
    \caption{Ablation study across different training steps. \ourmethod{} achieves the best aesthetics, semantic alignment, edit consistency, and realism throughout most training steps.}
    \label{fig:step_eval}
\end{figure*}

\begin{table}[htbp]
\centering
\caption{Ablation Study regarding the influence of different prompts for T2I rewards on SigLIP semantic consistency.}
\label{tab:ablation_prompt_eval}

\resizebox{\columnwidth}{!}{
\begin{tabular}{lccc}
\toprule

Method
& \shortstack[c]{Pure Edit\\Prompt}
& \shortstack[c]{Offline SOTA\\Prompt}
& \shortstack[c]{\ourmethod{}\\Prompt} \\

\midrule

w/o Alignment
& 0.0769
& 0.0768
& 0.0784 \\

\textbf{Lever} Pure Edit
& 0.0772
& 0.0766
& 0.0785 \\

\textbf{Lever} Pure VLM
& 0.0770
& 0.0765
& 0.0784 \\

\textbf{Lever} Offline SOTA
& 0.0767
& 0.0763
& 0.0782 \\

\textbf{\ourmethod{}}
& \textbf{\underline{0.0777}}
& \textbf{\underline{0.0770}}
& \textbf{\underline{0.0788}} \\

\bottomrule
\end{tabular}
}
\end{table}

\subsection{The Merits of Introducing T2I Rewards}
To verify our central hypothesis that the standard dimensions of editing quality can be mapped into the T2I reward space, we evaluate the transferred reward at two complementary levels. At the reward level, a successful transfer should correctly distinguish preferred edits from inferior ones, even without an editing-specific evaluator that jointly processes the reference image, instruction, and edited output. At the optimization level, it should provide a stable learning signal whose gains translate into better instruction execution, semantic and source consistency, and image quality. We therefore evaluate reward accuracy on annotated preference pairs, analyze reward trajectories during training, and compare downstream editing performance on Lever-Bench, EditReward-Bench \cite{luo2026editscore}, and GEdit \cite{liu2025step1x}. Together, these evaluations test not only whether a transferred T2I reward assigns meaningful preference scores, but also whether its preference knowledge can serve as effective supervision for image-editing RL.

Our reward-aligned target descriptions enable T2I rewards to provide a stronger editing-preference signal than directly applying them to raw edit instructions. At the reward level, we compare three variants on preference pairs with ground-truth annotations: HPSv3 conditioned on the reward-aligned target descriptions generated by \ourmethod{} (\emph{HPSv3+Ours}), HPSv3 conditioned directly on the raw edit instruction (\emph{HPSv3+Edit}), and the editing-specific reward EditScore. As shown in Fig.~\ref{fig:T2I_effect}(a), \emph{HPSv3+Ours} outperforms EditScore by approximately 9 percentage points in prompt-following accuracy. In contrast, directly conditioning HPSv3 on the edit instruction leads to a pronounced drop in consistency accuracy. This is because a relative edit instruction specifies what should change, but does not explicitly describe the source content that should be preserved. At the optimization level, the Forensic-Chat reward curves in Fig.~\ref{fig:T2I_effect}(b) show that \ourmethod{} consistently receives higher rewards than the alternatives throughout training. These results validate the central role of our target-description interface: converting the relative editing request into a self-contained target state makes mature T2I rewards better suited to evaluating image-editing preferences and, therefore, improves fine-tuning of the image-editing model.

\subsection{Effectiveness of Execution and Consistency}
Beyond improving the optimized T2I reward values, T2I-only supervision also yields clear gains on the core edit-aware dimensions of instruction execution and reference consistency.
Transferring T2I rewards substantially improves instruction execution while also providing strong semantic consistency. As shown in Tab.~\ref{tab:editing_metrics}, Lever-based methods on the EditScore Execution (prompt-following) metric consistently outperform the unaligned base image-editing model, indicating that the transferred reward provides an effective learning signal for executing the requested edit. Moreover, Lever-based methods show higher consistency, naturalness and execution on editscore. However, admittedly, a low-quality online target prompt may lead to a lower execution score, as the Editscore-Execution score of \textbf{Lever} Pure VLM. Lastly, in GEdit, \ourmethod{} achieves the best semantic score and the highest overall score. Since these reward models are also involved in optimization, we use the independent SigLIP consistency and GEdit as the extra evidence for consistency evaluation. As shown in Tab.~\ref{tab:editing_metrics}, \ourmethod{} has the best SigLip and GEdit S score (consistency) over all competitors. Therefore, the reward-aligned target descriptions generated by \ourmethod{} prove to provide a more effective semantic representation of the desired edited state than raw or generic textual conditions.

\subsection{Effectiveness of Image Quality}
Here, we show that
\ourmethod{} achieves the strongest alignment with the transferred T2I quality preferences on Lever-Bench. As reported in Tab.~\ref{tab:editing_metrics}, it achieves the highest HPSv3 score by a clear margin and the best RealGen score. The large HPSv3 gain is consistent with the objective of Stage~1: the learned target descriptions explicitly bridge the editing task and the preference space represented by the T2I reward. The corresponding RealGen improvement further shows that the resulting edited images align well with the complementary realism- and consistency-oriented reward signal.

\subsection{Ablation Study}
\ourmethod{} consistently outperforms the compared methods across different training stages regarding metrics such as HPSv3, SigLIP, Editscore consistency, and Realgen v1, as shown in Fig.~\ref{fig:step_eval}. To disentangle the effect of description quality from that of the edited images, we recompute SigLIP similarity using four alternative textual conditions: the raw edit instruction, descriptions from the vanilla VLM (\emph{Pure VLM}), descriptions generated by the strong offline model (\emph{Offline SOTA}), and the reward-aligned target descriptions generated by \ourmethod{}. The recomputed SigLIP scores are listed in Tab.~\ref{tab:ablation_prompt_eval}, which further reveals two clear observations. First, for the same edited image, the target descriptions generated by \ourmethod{} consistently achieve the highest SigLIP scores, indicating better semantic alignment between our descriptions and the intended edited content. Second, under the same textual description, images generated by \ourmethod{} also achieve the highest SigLIP scores, indicating that our edited images better match the intended target semantics. Together, these results demonstrate improvements in target-description quality and image-text semantic alignment.

\subsection{Qualitative Analysis}
Fig.~\ref{fig:exp3} presents qualitative comparisons from four complementary perspectives: execution (prompt-following), aesthetics, consistency, and naturalness. As shown in Fig.~\ref{fig:exp3}, for prompt-following, \ourmethod{} successfully transforms only the dog into a statue with marble texture, while the offline-based method loses consistency with the dog's appearance, and the Pure VLM method mistakenly maintains the original feather texture that contradicts marble property. Moreover, \ourmethod{} places the red sports car in the correct position, and is rendered with higher aesthetics and naturalness. Additionally, \ourmethod{} shows impressive aesthetics across global/local/drawing edits, and aligns `surboard' with the correct objects while maintaining human head pose, demonstrating high semantic and source consistency. Lastly, for the object removal task, \ourmethod{} recovers hidden leaves with higher naturalness quality.

%% file: chapter/Conclusion.tex
\section{Conclusion}

We show that the mature T2I reward ecosystem can be effectively transferred to image-editing RL without relying exclusively on editing-specific rewards. We introduce \ourmethod{}, a two-stage framework that bridges relative editing instructions and the self-contained target descriptions expected by T2I rewards by first learning a reward-aligned target-description policy and then freezing it to guide downstream image-editing optimization. Experiments on Lever-Bench and existing benchmarks show competitive edit alignment and source preservation, while outperforming intuitive transfer baselines based on raw edit instructions, generic VLM descriptions, and offline high-quality prompts. These show that effective image-editing RL does not necessarily require separately designed rewards for every editing objective. Instead, diverse visual preferences in T2I rewards can be reused through an appropriate target-description interface.